\pdfoutput=1

\documentclass[11pt]{article}

\usepackage[final]{acl}
\usepackage{todonotes}

\usepackage{times}
\usepackage{latexsym}

\usepackage[T1]{fontenc}

\usepackage[utf8]{inputenc}

\usepackage{microtype}

\usepackage{inconsolata}

\usepackage{graphicx}

\usepackage{wrapfig}
\usepackage{amssymb} 
\usepackage{pifont}  

\usepackage{hyperref}
\usepackage{url}
\usepackage{multirow}
\usepackage{amsmath}
\usepackage{booktabs}
\usepackage{indentfirst}
\usepackage{twemojis}
\definecolor{cred}{RGB}{172,45,57}
\usepackage{xspace}
\usepackage[capitalize,noabbrev,nameinlink]{cleveref}

\newcommand{\ourmethod}{{\color{cred} 
\texttwemoji{1f4af}-LongBench}\xspace}
\newcommand{\ourmethodNoColor}{{\texttwemoji{1f4af}-LongBench}\xspace}
\newcommand{\ourmetric}{\text{LongScore}\xspace}

\newcommand{\smallcircled}[1]{%
\tikz[baseline=(char.base)]{
    \node[shape=circle,draw,inner sep=0.2pt,minimum size=0.85em] (char) {\tiny#1};}} 
\usepackage{enumitem}

\title{\ourmethod: Are \textit{de facto} Long-Context Benchmarks Literally Evaluating Long-Context Ability?}

\author{
 \textbf{Wang Yang$^1$, Hongye Jin$^2$, Shaochen Zhong$^3$, Song Jiang$^4$, Qifan Wang$^5$}\\
 \textbf{Vipin Chaudhary$^1$, Xiaotian Han$^1$} \\
 $^1$Case Western Reserve University $^2$Texas A\&M University $^3$Rice University\\ $^4$University of California, Los Angeles $^5$Meta \\
 \texttt{\{wxy320,vipin,xhan\}@case.edu, jhy0410@tamu.edu, hz88@rice.edu 
} \\
 \texttt{
 songjiang@ucla.edu, wqfcr@meta.com
 }
 }
\begin{document}
\maketitle

\begin{abstract}

Long-context capability is considered one of the most important abilities of LLMs, as a truly long context-capable LLM shall enable its users to effortlessly process many originally exhausting tasks — e.g., digesting a long-form document to find answers v.s., directly asking an LLM about it. However, existing real-task-based long-context evaluation benchmarks have a few major shortcomings. For instance, some Needle-in-a-Haystack-like benchmarks are too synthetic, and therefore do not represent the real world usage of LLMs. While some real-task-based benchmarks like LongBench avoid this problem, such benchmarks are often formed in a way where each data sample has a fixed sequence length, which not only makes them solely suitable for models with a certain range of context windows, but also lacks a proxy to know at what length the model/method-of-interest would fail. Last, most benchmarks tend to not provide proper metrics to separate long-context performance from the model's baseline ability, so when conducting a cross-model/recipe comparison, such conflation makes the user unable to understand how exactly one model or recipe excels at the long-context task in relation to its baseline ability. \quad To address these issues, we introduce a length-controllable, real-life reflective benchmark with a novel metric that disentangles baseline knowledge from long-context capabilities. Experiments demonstrate the superiority of our datasets in effectively evaluating LLMs. All assets are available at \url{https://github.com/uservan/100-LongBench.git}.

\end{abstract}

\begin{figure*}[t]
  \includegraphics[width=1\linewidth]{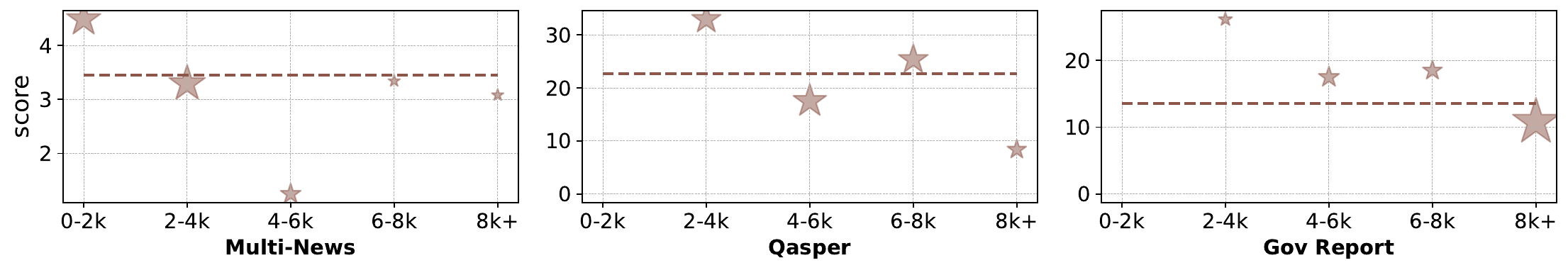}\vspace{-10pt}
  \caption {Illustration of LM-Infinite~\cite{han2024lm}, a long-context enhancement method's performances on three LongBench tasks. The colored dashed lines represent the average score of each model on the corresponding task. The size of the markers corresponds to the proportion of each text length within the entire dataset. The larger the marker, the higher the proportion. The results exhibit significant variation across tasks of different lengths within the same dataset. More results of other methods are in \cref{Results of models’ long-text enhancement methods on Longbench}.}

\label{fig:different_length}
\end{figure*}

\section{Introduction}

\begin{table}
  \centering
  \caption{
    Models' ranking on Ruler~\cite{hsieh2024ruler} with different metrics. \textbf{Base Ability} represents model’s score within $4k$ context. \textbf{Old/Proposed Metric} represents the average score across various lengths using traditional metric/our proposed metric. $96.5_{(1)}$ indicates a score of 96.5 with a rank of 1. More details are in \cref{table:ruler results on metrics2}. Comparing the ranking of Old Metric and Proposed Metric reveals that the rankings of the old metrics are heavily influenced by the model's inherent abilities, which might not really reflect long-context ability.
  }\vspace{-10pt}
  \resizebox{\columnwidth}{!}{
  \begin{tabular}{l|c|cc}
     \toprule
  \multirow{2}{6em}{\centering  \textbf{Model}~(size,length)} & 
  \multirow{2}{3em}{\centering Base Ability} & \multirow{2}{4em}{\centering Old Metric}& \multirow{2}{5em}{\centering Proposed Metric} \\
    & & \\
    \midrule
    Llama3.1 (70B, 128K) & $96.5_{\color{red}(1)}$
    & $88.2_{\color{red}(1)}$ & $-8.6_{\color{orange}(2)}$ \\
    Yi (34B, 200K)~\cite{young2024yi}  & $93.3_{\color{orange}(2)}$
    & $86.3_{\color{orange}(2)}$ & $-7.5_{\color{red}(1)}$ \\
    Phi3-medium (14B, 128K) & $93.3_{\color{blue}(3)}$
    & $79.1_{\color{blue}(3)}$ & $-15.1_{\color{green}(4)}$ \\
    LWM (7B, 1M)~\cite{liu2024world}  & $82.3_{\color{green}(4)}$
    & $70.8_{\color{green}(4)}$ & $-13.9_{\color{blue}(3)}$ \\
   \bottomrule
  \end{tabular}
  }
  \label{table:ruler results on metrics}
  \vspace{-2em}
\end{table}

The long-context capability has become one of the fundamental competencies~\cite{gao2024train,liu2024lost,li2024chain, agarwal2024many} of contemporary large language models (LLMs) because it takes the average human critical time and effort to digest long-form context, making a long-context-capable LLM beyond desirable. To assess the long-context capabilities of LLMs, various evaluation benchmarks and metrics have been proposed, including LongBench~\cite{bai2023longbench}, L-Eval~\cite{an2023leval}, NIAH (Needle in the Haystack), RULER~\cite{hsieh2024ruler}, Ada-LEval~\cite{wang2024adaleval} and Loogle~\cite{fli2023loogle}. However, these benchmarks often exhibit at least one of the following three shortcomings:

(1) They rely on purely \textbf{synthetically-constructed content that is not real-life reflective}. Synthetic benchmarks such as \textsc{NIAH} or Passkey Retrieval often demand the retrieval of a source (e.g., a string of digits or a phrase) that bears no semantic or task relevance to the padding content (e.g., unrelated blog posts). This kind of highly artificial task does not properly reflect how LLMs are utilized in typical real-world settings, where information of similar nature is often joined together for a reader to understand and digest.

(2) They adopt a \textbf{fixed input length per data sample}, making them suitable only for certain LLMs with compatible context windows. This is a major problem because context windows have grown significantly, thanks to the development of context extension techniques and post-training recipes. With Llama 3.1~\cite{dubey2024llama} claiming a context window of 128k (in contrast to the 4k context of Llama 2), many once ``long-context'' datasets have already become outdated. It is therefore foreseeable that many evaluations we see today will no longer be reflective as time passes.\quad Moreover, having different lengths per individual data sample makes the evaluation reading unintuitive in several ways: E.g., for model evaluation, it is hard to tell at what length it would fail or prevail, because we only get the aggregated reading upon questions of different lengths. For method evaluation, many constant-budget compression works — like StreamingLLM~\cite{xiao2023streamingllm} and InfLLM~\cite{xiao2024infllm} — have an arbitrarily set constant budget that is applied to all inputs, ignoring the fact that this budget may exceed certain data samples. As a result, the reported ``compressed performance'' often turns into an unknown mixture of both compressed and uncompressed results, complicating the transparency of assessments.

(3) They do not address the \textbf{conflation between base ability and long-context capability}, as these benchmarks evaluate long-context capabilities solely based on long-context tasks without isolating the influence of a model’s baseline abilities. Thus, some readings can be tricky to digest when factors cannot be perfectly ablated. For instance, suppose we have two different base models, each has undergone their own continuous pretraining recipes for context extension (e.g., Llama and Qwen), \textit{which extension recipe is likely better?} Applying both recipes to the same base model for direct comparison is often impractical due to compute and dataset resource limitations. Naturally, one avenue of evaluation is to measure the long context performance relative to the short context performance for an educated understanding, but such kind of measurements is largely missing in most existing long-context benchmarks.

In this work, we attempt to alleviate such problems by providing a \textbf{new benchmark} involving a rich set of length-controllable real-life-reflective tasks — \ourmethodNoColor — and a \textbf{new evaluation metric} — LongScore — which leads to significant shifts in model rankings compared to traditional performance-based evaluations, as shown in \cref{table:ruler results on metrics}. We first validate the reliability of the proposed \ourmethodNoColor and the effectiveness of \ourmetric. We then \textbf{comprehensively benchmark} various open-source models, providing \textbf{fresh insights} into long-context evaluation and offering a more accurate assessment that better reflects models’ true abilities to handle extended contexts.

\section{Motivation: why do we need to refine long-context benchmarks?} 

 
\paragraph{Performance variance across task lengths} Evidenced by \cref{fig:different_length}, the performance of LM-Infinite exhibits significant variation across tasks of different lengths within the same dataset. Many long-context datasets have uneven length distributions, introducing biases in evaluating a model’s long-context capability. To validate this hypothesis, we train models using five different long-context enhancement methods and evaluate their performances across varying lengths on the LongBench dataset. From \cref{fig:different_length}, we observe the following: (1) Performance Variation: All five models demonstrate performance differences across different text lengths within the same dataset. (2) Alignment with Dominant Lengths: A model’s average performance aligns closely with its performance on the length range with the highest proportion of samples. For instance, on Multi-News dataset, each model's average performance is close to its performance on samples within the 0–4k length range, which represents the largest share of the dataset. 
These findings highlight the need for length-aware evaluations of long-context capabilities. A more robust approach involves testing model performance on  $N$ samples across diverse lengths to obtain a comprehensive assessment of its long-context capability. More results of other methods are in \cref{Results of models’ long-text enhancement methods on Longbench}.

\begin{figure}[t]
  \includegraphics[width=\columnwidth]{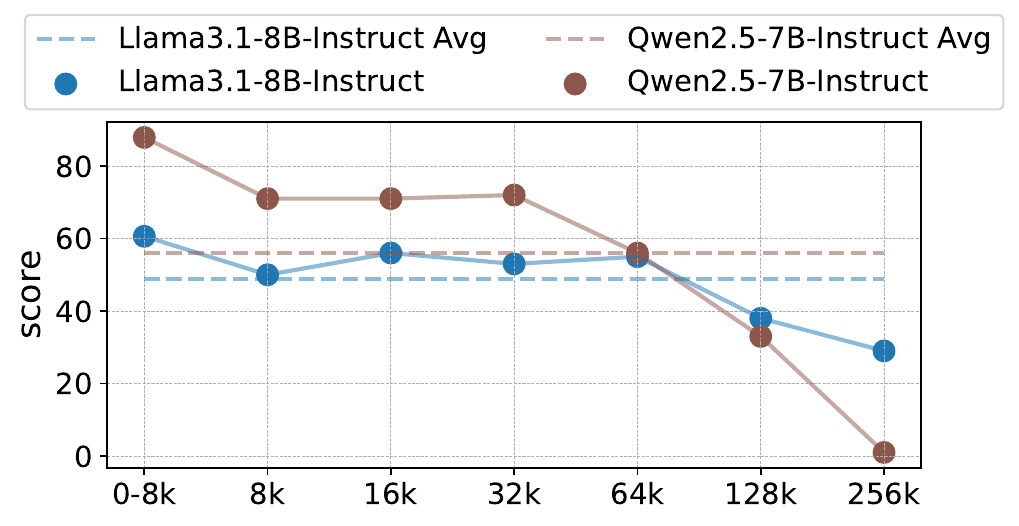}
  \caption{Comparison between LLaMA 3.1-8B-Instruct and Qwen 2.5-7B-Instruct on the Counting Star task across varying text lengths. The dashed line represents the average score across all context lengths. LLaMA 3.1-8B-Instruct performs worse than Qwen 2.5-7B-Instruct on short texts but excels on extremely long texts, indicating its superior long-context extension capability.}\label{fig:changes}
\end{figure}

\paragraph{Ineffectiveness of current metrics for evaluating long-context capability} Evidenced by \cref{fig:changes}, existing long-context metrics primarily rely on the average performance across the benchmark. However, this approach can be misleading as it conflates the model's inherent task-specific ability with its pure long-context capability. As illustrated in \cref{fig:changes}, LLaMA 3.1-8B-Instruct performs worse than Qwen 2.5-7B-Instruct on short texts but excels on extremely long texts, such as $128k$ and $255k$, indicating its superior long-context extension capability. In this task, the average performance suggests that Qwen 2.5-7B-Instruct is the better model. But a closer inspection reveals that LLaMA 3.1-8B-Instruct has a distinct advantage in handling extremely long texts, despite its weaker performance on shorter inputs. This discrepancy underscores the need to separate a model’s base ability (on short texts) from its long-context capability. To address this, we propose a novel metric that accurately captures a model’s true 
capability to handle long contexts from Base Ability.

\section{How to truly evaluate Language Models' long-context capability?}
To address the two identified problems, we 1) construct a length-controllable long-context benchmark to reduce performance variance across task lengths, and 2) introduce LongScore, a new metric designed to accurately evaluate long-context capabilities by disentangling the model's baseline abilities.
In detail, we restructure the long-context datasets, based on LongBench, L-EVAL, and other benchmarks. We then design a new pipeline to generate controllable-length long contexts by combining different articles. Additionally, we introduce a filtering mechanism in QA-related tasks to mitigate prior knowledge. Subsequently, We propose a new metric to isolate a model’s long-text capability from Base Ability (performance on short texts).

\begin{table*}
  \centering
  \caption{
  Details of dataset construction for each task. To generate a context of a specified length like $128k$, we randomly select multiple articles from the Noisy Context Source datasets as distractor articles. A single article is randomly chosen from Real Context Source datasets as the ground truth article. Distractor articles and the ground truth article are combined to form the whole context, ensuring that the whole context length is less than 128k and the order of all articles is shuffled. The bottom of the table contains different datasets from other benchmarks. N/A indicates that the task does not require Context Sources because the questions are synthetic rather than derived from a dataset. More details about how to construct each task are in \cref{sec:construction}.
  }\vspace{-10pt}
  \begin{tabular}{lccc}
    \hline
     \textbf{Task Name}  & \textbf{Real Context Sources}
    & \textbf{Noisy Context Sources} & \textbf{Evaluation Metric} \\
    \hline
    KV Retrieval      & N/A          & \smallcircled{1} \smallcircled{2} \smallcircled{3} \smallcircled{4}  \smallcircled{5} \smallcircled{6} \smallcircled{7}  \smallcircled{8}  \smallcircled{9}  &   Accuracy   \\
    Counting Stars    & N/A         & \smallcircled{1} \smallcircled{2} \smallcircled{3} \smallcircled{4}  \smallcircled{5} \smallcircled{6} \smallcircled{7}  \smallcircled{8}  \smallcircled{9}  &   Accuracy   \\
    \hline
    Passage Retrieval & \smallcircled{9} \smallcircled{10}  \smallcircled{11} \smallcircled{12}  \smallcircled{13} \smallcircled{14} \smallcircled{15}  &  \smallcircled{9} \smallcircled{10}  \smallcircled{11} \smallcircled{12}  \smallcircled{13} \smallcircled{14} \smallcircled{15}     &     Accuracy  \\

    Passage Count   &                   \smallcircled{1} \smallcircled{2} \smallcircled{3} \smallcircled{4}  \smallcircled{5} \smallcircled{6} \smallcircled{7}  \smallcircled{8}  \smallcircled{9}     & N/A     &   Accuracy   \\
    \hline
    Single-doc QA     &    
    \smallcircled{1} \smallcircled{2} \smallcircled{3} \smallcircled{4}  \smallcircled{5} \smallcircled{6} \smallcircled{7}  \smallcircled{8}
    & \smallcircled{1} \smallcircled{2} \smallcircled{3} \smallcircled{4}  \smallcircled{5} \smallcircled{6} \smallcircled{7}  \smallcircled{8}
    &   LLM-based Metric   \\
    Multi-doc QA      &         
    \smallcircled{16} \smallcircled{17} \smallcircled{18} \smallcircled{19}
    & \smallcircled{1} \smallcircled{2} \smallcircled{3} \smallcircled{4}  \smallcircled{5} \smallcircled{6} \smallcircled{7}  \smallcircled{8}
    &   LLM-based Metric   \\
    \hline
    Single-doc Sum     &   
    \smallcircled{1} \smallcircled{11} \smallcircled{12} \smallcircled{13} \smallcircled{14} \smallcircled{15}
    & \smallcircled{1} \smallcircled{11} \smallcircled{12} \smallcircled{13} \smallcircled{14} \smallcircled{15}
    & LLM-based Metric     \\
    Multi-doc Sum     &       
    \smallcircled{20}
    & \smallcircled{1} \smallcircled{11} \smallcircled{12} \smallcircled{13} \smallcircled{14} \smallcircled{15}  
    & LLM-based Metric   \\
    \hline
    \hline
    \multicolumn{4}{|c|}{
    \smallcircled{1} qasper 
    \smallcircled{2} multifieldqa\_en  \smallcircled{3} narrativeqa 
    \smallcircled{4} multidoc\_qa 
    \smallcircled{5} legal\_contract\_qa  
    } \\
    \multicolumn{4}{|c|}{
    \smallcircled{6} financial\_qa
    \smallcircled{7} natural\_question \smallcircled{8} scientific\_qa \smallcircled{9} cnn\_dailymail \smallcircled{10} gov\_report
    } \\
    \multicolumn{4}{|c|}{
    \smallcircled{11} qmsum 
    \smallcircled{12} patent\_summ 
     \smallcircled{13} tv\_show\_summ 
     \smallcircled{14} review\_summ 
     \smallcircled{15} meeting\_summ
    }\\
    \multicolumn{4}{|c|}{
    \smallcircled{16}hotpotqa 
     \smallcircled{17}2wikimqa 
     \smallcircled{18}musique 
     \smallcircled{19}rag-mini-bioasq
     \smallcircled{20} multi\_news\_e
    }\\
    \hline
    
  \end{tabular}

  \label{table: task sources}
\end{table*}

\subsection{Construct a new long-context benchmark}

We categorize tasks into four types, each consisting of two tasks with different levels of difficulty, resulting in a total of eight tasks. The types and their corresponding tasks are: \textbf{\textit{Key Retrieval (including KV Retrieval and Counting Stars), Information Retrieval (including Passage Retrieval and Passage Count ) , Information Comprehension (including Single-doc QA and Multi-doc QA) and Information Summarization (including Single-doc Sum and Multi-doc Sum) }}. \cref{table: task sources} provides details for each task, including: Real Context Sources(the original context of the question used in the task), Noisy Context Sources(the source of additional context that may contain irrelevant or distracting information) and Evaluation Metric(the metric used to assess model performance for each task). All of these datasets are from other benchmarks 
like LongBench, etc. Detailed information on context construction, question setup, and evaluation metrics, are in \cref{sec:construction}.

\begin{figure}[t]
\includegraphics[width=\columnwidth]{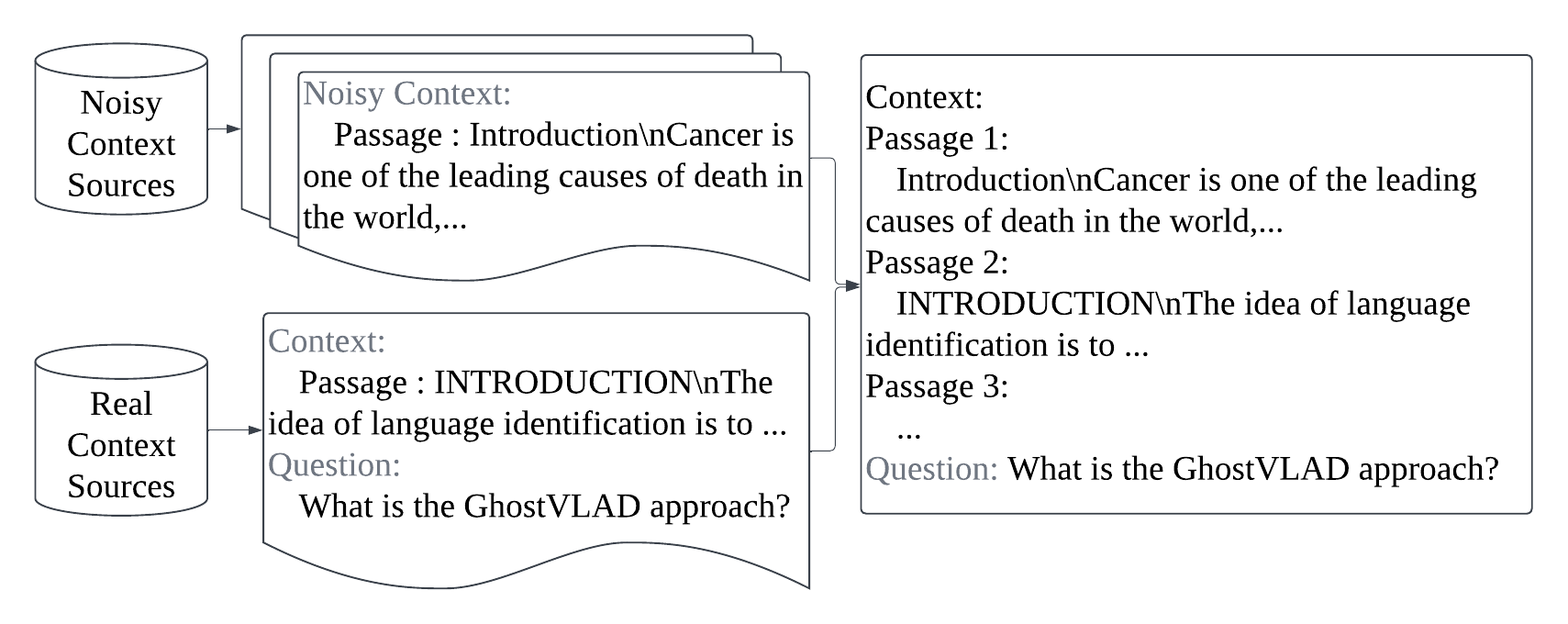}
  \caption{Illustration of the Data Generation Process for the Single-Doc QA Task}\vspace{-10pt}
\label{fig:generate}
\end{figure}

\textbf{How to generate a controllable-length context?} In \ourmethodNoColor, the context for each task is controllable, such as generating a context of approximately $128k$ tokens. To achieve this, we first randomly select one article from Real Context Sources as the ground truth article. Then, we randomly sample a number of articles from Noisy Context Sources as distractor articles. These distractor articles are combined with the ground truth article to construct the whole context, ensuring that the total context length is close to but less than $128k$. Finally, the order of all articles is shuffled to create the context. \cref{fig:generate} illustrates the data generation process for Single-Doc QA task, showing how questions, answers, and contexts are prepared.

\begin{figure}[t]
  \includegraphics[width=\columnwidth]{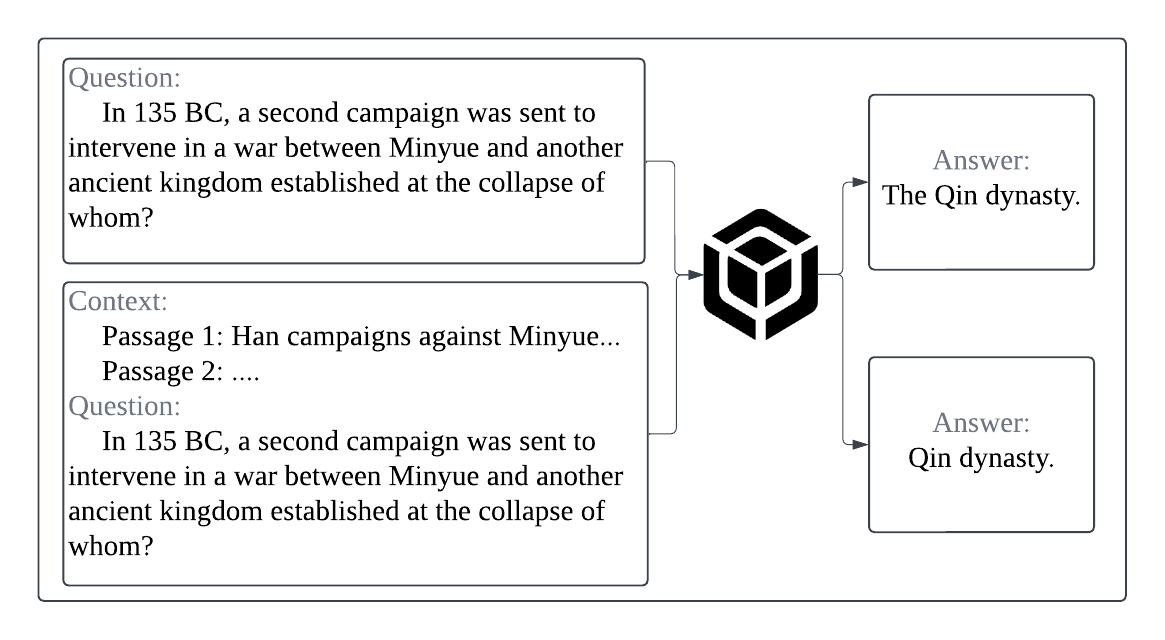}\vspace{-10pt}
  \caption{One sample in Question Answering where models provide accurate answers regardless of context}
  \label{fig:no_context_answer}
\end{figure}

\textbf{QA Filtering Mechanism}. For Multi-Doc QA and Single-Doc QA tasks, we introduce a filtering mechanism to eliminate the influence of the model’s inherent prior knowledge. When evaluating a model’s long-context capabilities, prior knowledge is often overlooked. For instance, in question-answering (QA) tasks, the model might memorize the answers to certain questions during pretraining. As shown in \cref{fig:no_context_answer}, the model accurately answer questions based on its prior knowledge even without any contexts. In such cases, the model’s response is not derived from the provided context but from its memorized knowledge. This oversight can lead to inflated performance metrics, misrepresenting the model’s actual ability to process and comprehend long contexts. To filter out the model’s prior knowledge, we introduce a QA filtering mechanism. In a no-context scenario, if the model’s response score exceeds a certain threshold, it indicates that the model is relying on prior knowledge, showing the data should be excluded.

Although our length-controlled datasets are synthetically constructed, they are carefully designed to better reflect real-world usage scenarios, which we called as real-life reflective. Specifically, each instance is composed by selecting a task-relevant example as the \textit{source} (e.g., a summarization prompt and document), and padding it with additional samples that belong to the same domain or task type (e.g., other documents suitable for summarization). This construction ensures that all components of the input are contextually aligned and task-compatible, mimicking common usage patterns in long-context settings, such as concatenated inputs in retrieval-augmented generation pipelines.

\subsection{\ourmetric: a new long-context metric}

\begin{table*}
\centering

\caption{Comparison of long-context benchmarks: Longbench~\cite{bai2023longbench}, L-Eval~\cite{an2023leval}, $\infty$-Bench~\cite{zhang2024infintebench}, NIAH (Needle In A Haystack), RULER~\cite{hsieh2024ruler}, Helmet~\cite{yen2024helmet}, and our \ourmethodNoColor. L: input tokens. Controllable: The benchmark can generate contexts of specified lengths. Diverse Tasks: The tasks are varied and not limited to a single type. LLM-based Metric: Metrics in some tasks are designed based on large language models for more accurate evaluation. LC Distinction: Effectively separates the model’s base ability from its long-text capability. QA Filter: Implements measures to remove the influence of the model’s prior knowledge in QA tasks. The tasks in NIAH and RULER are synthetic, so they do not require LLM-based metrics or QA filtering.}\vspace{-10pt}
{\small
\begin{tabular}{ccccccc}
\toprule
\multirow{2}{4em}{\centering Dataset} & \multirow{2}{6em}{\centering $L > 128k$} & \multirow{2}{6em}{\centering Controllable} & \multirow{2}{6em}{\centering Diverse Tasks} & \multirow{2}{6em}{\centering LLM-based Metric} & \multirow{2}{6em}{\centering LC distinction} & \multirow{2}{6em}{\centering QA Filter}  
\\ \\ 
\midrule
Longbench & \textcolor{red}{\ding{55}}  & \textcolor{red}{\ding{55}} & \textcolor{green}{\checkmark} & \textcolor{red}{\ding{55}} & \textcolor{red}{\ding{55}} & \textcolor{red}{\ding{55}} \\
L-EVal & \textcolor{red}{\ding{55}} & \textcolor{red}{\ding{55}} & \textcolor{green}{\checkmark} & \textcolor{green}{\checkmark} & \textcolor{red}{\ding{55}}  & \textcolor{red}{\ding{55}} \\
$\infty$-Bench & \textcolor{green}{\checkmark} & \textcolor{red}{\ding{55}} & \textcolor{green}{\checkmark} & \textcolor{red}{\ding{55}} & \textcolor{red}{\ding{55}} & \textcolor{red}{\ding{55}} \\
NIAH & \textcolor{green}{\checkmark} & \textcolor{green}{\checkmark} & \textcolor{red}{\ding{55}} &  & \textcolor{red}{\ding{55}} & \\
RULER & \textcolor{green}{\checkmark} & \textcolor{green}{\checkmark} & \textcolor{green}{\checkmark} &  & \textcolor{red}{\ding{55}} &  \\
Helmet & \textcolor{green}{\checkmark} & \textcolor{green}{\checkmark} & \textcolor{green}{\checkmark} & \textcolor{green}{\checkmark} & \textcolor{red}{\ding{55}} & \textcolor{red}{\ding{55}} \\
\midrule
\ourmethodNoColor & \textcolor{green}{\checkmark} & \textcolor{green}{\checkmark} & \textcolor{green}{\checkmark} & \textcolor{green}{\checkmark} & \textcolor{green}{\checkmark} &  \textcolor{green}{\checkmark}  \\
\bottomrule
\end{tabular}
}
\label{table: comparsion to other benchmarks}
\end{table*}

As illustrated in \cref{fig:changes}, directly using a model’s scores across various text lengths to assess its long-context capability introduces inherent biases. To address this limitation, we propose a new metric that disentangles the model’s base ability from its long-context capability, allowing for a more accurate and comprehensive evaluation.

\textbf{Base Ability}. It refers to the model’s score when conducting short-context tasks. To estimate Base Ability, we sample  $N$  instances from short text lengths (like $2k$, $4k$, $6k$). For each length,  $N/3$ samples are selected, and the model’s average score across these lengths is computed:

\begin{equation}
  \label{eq:Base Ability}
 \text{Base Ability} = \frac{S_{2k} + S_{4k} + S_{6k}}{3}
\end{equation}
where $S_{*k}$ represents the performance of model with the ${*-k}$ length.

\begin{figure*}[t]
\includegraphics[width=1\linewidth]{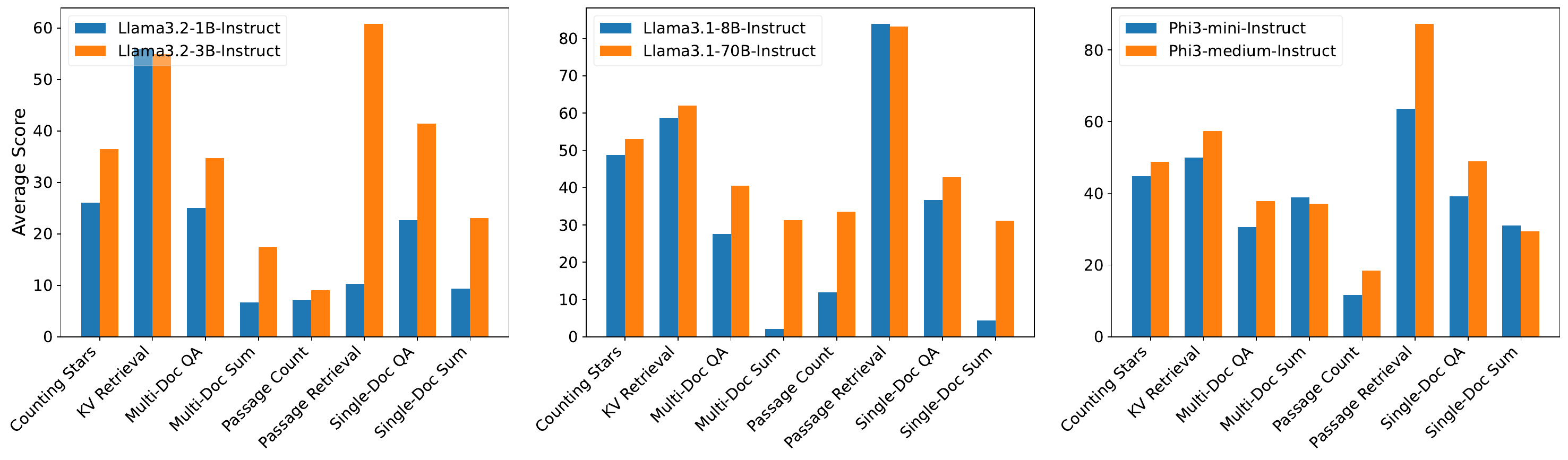}\vspace{-10pt}
  \caption {Verification of the reliability of \ourmethodNoColor: results of two models of different sizes from the same LM family tree, showcasing their average scores in different tasks. These findings confirm a well-established trend: within the same series, larger models generally outperform smaller ones, reinforcing the reliability of \ourmethodNoColor.}
 \label{fig: verification}
\end{figure*}

\textbf{\ourmetric} ($\text{LC}_l$) is our proposed metric. For longer lengths (e.g., 8k, 16k, 32k), we calculate the score on $N$ instances for each length.  $\text{LC}_l$ at a given length $l$ is then defined as:

\begin{equation}
  \label{eq:LTC}
 \text{LC}_l = \frac{S_l - \text{Base Ability}}{\text{Base Ability}}
\end{equation}

\ourmetric separates the model’s Base Ability from Long-context Capability. Our metric focuses on the relative improvement or decline at longer lengths and provides a more precise assessment of long-context capabilities without being influenced by the model’s Base Ability. It enables consistent and unbiased comparisons of long-context capabilities across different models and datasets.

\subsection{Compare to other benchmarks}

This section compares other long-context benchmarks with \ourmethodNoColor, highlighting their similarities and differences. The overall distinctions between benchmarks are presented in \cref{table: comparsion to other benchmarks}.
\begin{itemize}[leftmargin=0.5cm, itemindent=.0cm, itemsep=0.0cm, topsep=0.0cm]
    \item LongBench~\citep{bai2023longbench} is an early benchmark used to evaluate the long-context capabilities. It was the first to use a variety of tasks for evaluation, but the context length is generally limited to around $8k$, and the length distribution is uneven. As many current LLMs support context lengths of 128k and beyond, these benchmarks are no longer suitable.
    \item $\infty$-Bench~\cite{zhang2024infintebench} and L-Eval~\cite{an2023leval} are an improvement over benchmarks like LongBench, increasing the data length to over $128k$. However, the context length is not controllable, which limits its ability to comprehensively evaluate LLMs.
    \item NIAH and RULER~\citep{hsieh2024ruler} are designed with controllable context lengths and can control the position of the answer, specifically for evaluating long-context capabilities. These benchmarks are currently the leading tools to assess the long-context capabilities of LLMs.
    \item Helmet~\cite{yen2024helmet} is a newly proposed benchmark that not only allows for controllable context lengths but also designs a wide variety of tasks. It introduces the use of LLM-based metrics, providing a more refined way to evaluate long-context capabilities.
    \item \ourmethodNoColor generates controllable context-length tasks. Additionally, it introduces a new metric to distinguish between a model’s base ability and long-context capability, offering a more comprehensive and novel approach to evaluating long-context capabilities.
  
\end{itemize}

\section{Experimental Analysis}
In this section, we conduct comprehensive experiments to first validate the reliability of \ourmethodNoColor and the effectiveness of the proposed metric. They are then used to evaluate the long-context capabilities of several open-source models.

\begin{table}
  \centering
  \caption{
    Results of the average performance of five models across all tasks on \ourmethodNoColor. Base Ability represents the model’s score within lengths of $2k$, $4k$ and $6k$ \textbf{$Avg$ score} represents the average of score across lengths including $8k$, $16k$, $32k$, $64k$ and $128k$. \textbf{$Avg$ LC} represents the average of score by using our proposed metric, \ourmetric. $59.1_{(1)}$ indicates that the current model has a score of 59.1 at the given context length, with a ranking of 1. Claimed Length refers to the maximum context length that the model claims to support. Qwen 2.5-14B and Qwen 2.5-7B use YaRN to extend their context length to 128k. The original context length is specified in Claimed Length. 
    }\vspace{-8pt}
  \resizebox{\columnwidth}{!}{
  \begin{tabular}{c|c|c|cc}
     \toprule
  \multirow{2}{1.5em}{\centering  \textbf{Model}} & \multirow{2}{4em}{\centering Claimed Length} & \multirow{2}{3em}{\centering  Base Ability} & \multirow{2}{2em}{\centering $Avg$ \textbf{socre}}& \multirow{2}{2em}{\centering $Avg$ \textbf{LC}} \\
    \\
    \midrule
    Qwen2.5-14B-Instruct &  32K & $59.1_{(1)}$
    & $40.7_{(1)}$ & $-31.1_{(4)}$ \\
    Qwen2.5-7B-Instruct &  32K  & $57.4_{(2)}$
    & $39.8_{(2)}$ & $-30.6_{(3)}$ \\
    Llama3.1-8B-Instruct & 128K & $44.0_{(3)}$
    & $36.3_{(3)}$ & $-17.4_{(1)}$ \\
    Llama3.2-1B-Instruct  & 128K   & $28.7_{(4)}$
    & $20.4 _{(4)}$ & $-28.8_{(2)}$ \\
   \bottomrule
  \end{tabular}
  }
  \label{table:our results on metrics}
  \vspace{-1em}
\end{table}

\begin{figure}[t]
\centering
\includegraphics[width=0.9\columnwidth]{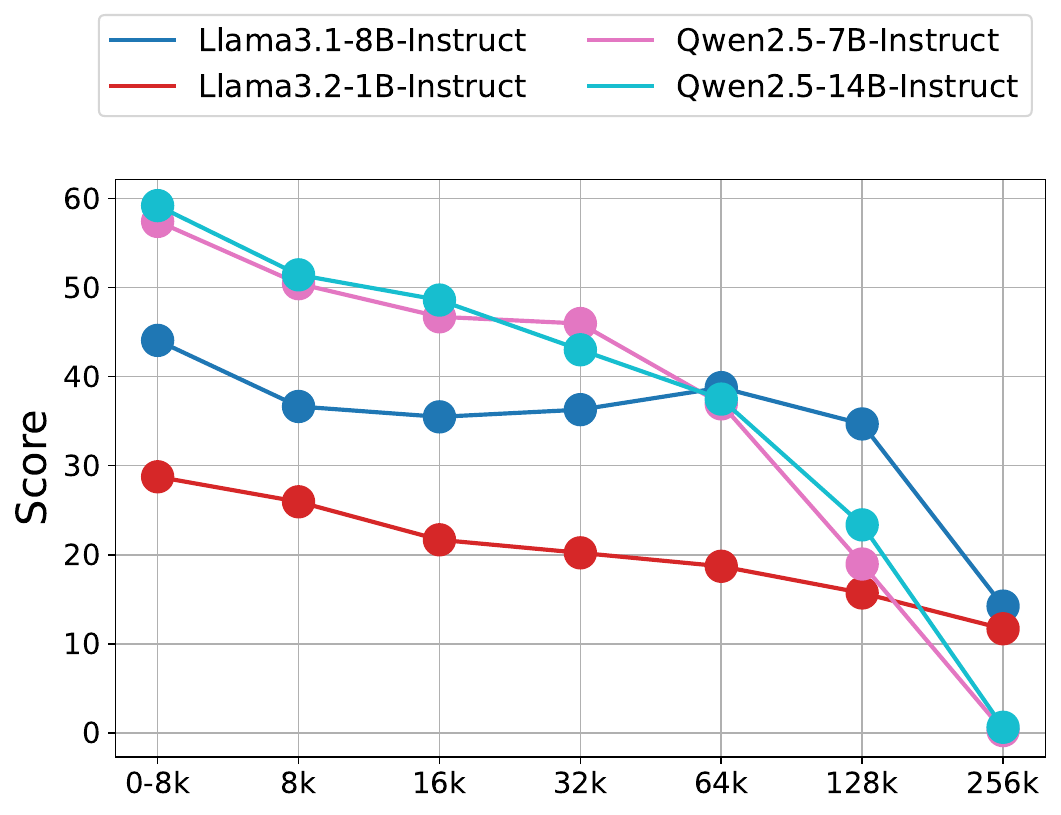}\vspace{-8pt}
  \caption{Results of four open-source models on all tasks in \ourmethodNoColor, showing their average scores of all eight tasks at different context lengths. }
  \label{fig:results original}
  \vspace{-1em}
\end{figure}

\begin{table*}
  \centering
  \caption{
    Results of 4 models' ranking in Ruler\cite{hsieh2024ruler} on different metrics. \textbf{Base Ability} represents the model’s score with a 4k-length context. $Avg$ represents the average of scores excluding the base score. $95.8_{(1)}$ indicates that the current model has a score of 95.8 at the given context length, with a ranking of 1. LC represents the score by our proposed metric, \ourmetric.  
  } \vspace{-10pt}
  \resizebox{\textwidth}{!}{
  \begin{tabular}{c|c|c|cc|cc|cc|cc|cc|cc}
     \toprule
  \multirow{2}{2em}{\centering  \textbf{Models}} & \multirow{2}{4em}{\centering Claimed Length} & \multirow{2}{3em}{\centering  \textbf{Base Ability}} & \multicolumn{2}{c} {\textbf{$8k$}} & \multicolumn{2}{c}{\textbf{$16k$}} & \multicolumn{2}{c}{\textbf{$32k$}} & \multicolumn{2}{c}{\textbf{$64k$}} & \multicolumn{2}{c}{\textbf{$128k$}} & \multicolumn{2}{c}{\textbf{$Avg$}} \\
      &   & & \textbf{score} & \textbf{LC} &  \textbf{score} & \textbf{LC} &  \textbf{score} & \textbf{LC} &  \textbf{score} & \textbf{LC} &  \textbf{score} & \textbf{LC} &  \textbf{score} & \textbf{LC} \\
    \midrule
    Llama3.1 (70B) &  128K & $96.5_{(1)}$
    & $95.8_{(1)}$ & $-0.7_{(2)}$
    & $95.4_{(1)}$ & $-1.1_{(1)}$
    & $94.8_{(1)}$ & $-1.7_{(1)}$
    & $88.4_{(1)}$ & $-8.3_{(1)}$
    & $66.6_{(2)}$ & $-30.9_{(3)}$ 
    & $88.2_{(1)}$ & $-8.6_{(2)}$ \\
    Yi (34B~\cite{young2024yi}) &  200K  & $93.3_{(2)}$
    & $92.2_{(3)}$ & $-1.1_{(3)}$
    & $91.3_{(2)}$ & $-2.1_{(2)}$
    & $87.5_{(2)}$ & $-6.2_{(2)}$
    & $83.2_{(2)}$ & $-10.8_{(2)}$
    & $77.3_{(1)}$ & $-17.1_{(1)}$ 
    & $86.3_{(2)}$ & $-7.5_{(1)}$ \\
    Phi3-medium (14B) & 128K & $93.3_{(3)}$
    & $93.2_{(2)}$ & $-0.1_{(1)}$
    & $91.1_{(2)}$ & $-2.3_{(3)}$
    & $86.8_{(3)}$ & $-6.9_{(3)}$
    & $78.6_{(3)}$ & $-15.7_{(3)}$
    & $46.1_{(4)}$ & $-50.5_{(4)}$  
    & $79.1_{(3)}$ & $-15.1_{(4)}$ \\
    LWM (7B)~\cite{liu2024world}  & 1M  & $82.3_{(4)}$
    & $78.4_{(4)}$ & $-4.70_{(4)}$
    & $73.7_{(4)}$ & $-10.4_{(4)}$
    & $69.1_{(4)}$ & $-16.0_{(4)}$
    & $68.1_{(4)}$ & $-17.2_{(4)}$
    & $65.0_{(3)}$ & $-21.0_{(2)}$
    & $70.8 _{(4)}$ & $-13.9_{(3)}$ \\
   \bottomrule
  \end{tabular}
  }
  \label{table:ruler results on metrics2}
\end{table*}

\begin{table*}[t]
  \centering
  \caption{
  \textbf{Comparison of models and methods under our proposed LongScore metric.}
  We present three evaluations to validate the discriminative power of LongScore: 
  (1) NTK vs. PI on 100-LongBench; 
  (2) LLaMA3-8B with different RoPE theta ratios; 
  (3) Gemini-1.5 variants from the HEMLET benchmark.
  In all cases, LongScore reflects performance differences that align with common understanding (e.g., NTK > PI, Gemini-Pro > Gemini-Flash), while amplifying meaningful gaps that are not visible with raw accuracy. The results highlight the discriminative ability and effectiveness of our proposed benchmark and metric.
  }\vspace{-8pt}
  \resizebox{\textwidth}{!}{
  \begin{tabular}{c|c|cccccc|cc}
     \toprule
     \textbf{Benchmark} & \textbf{Model / Method} & \textbf{base} & \textbf{8k} & \textbf{16k} & \textbf{24k / 32k} & \textbf{48k / 64k} & \textbf{128k / 256k} & \textbf{avg(score)} & \textbf{avg(\textsc{LongScore})} \\
     \midrule
     \multirow{2}{*}{100-LongBench} 
     & PI  & 19.18 & 16.47 & 17.67 & 17.10 & 17.67 & 0.44  & 13.87 & -27.68 \\
     & NTK & 19.39 & 15.72 & 16.53 & 16.70 & 17.17 & 12.88 & 15.83 & -18.40 \\
     \midrule
     \multirow{2}{*}{100-LongBench} 
     & LLaMA3-8B (ratio=1)   & 35.37 & 37.08 & 1.45 & 1.87 & 0.52 & 0.99  & 7.13 & -79.84 \\
     & LLaMA3-8B (ratio=64)  & 32.52 & 31.94 & 25.34 & 26.08 & 26.94 & 1.63 & 18.83 & -42.12 \\
     \midrule
     \multirow{2}{*}{HEMLET} 
     & Gemini-1.5-Flash & 59.6 & -- & 60.2 & 58.1 & 55.0 & 50.7 & 56.00 & -6.04 \\
     & Gemini-1.5-Pro   & 59.5 & -- & 60.1 & 59.9 & 57.0 & 54.1 & 57.77 & -2.90 \\
     \bottomrule
  \end{tabular}
  }
  \label{table:verification ntk pi}
\end{table*}

\subsection{Verification of the reliability of the proposed benchmark}

To verify the reliability of \ourmethodNoColor, we evaluate three model families (Llama 3.2, Llama 3.1, and Phi 3), selecting two different model sizes from each family. Since these are models of different sizes within the same series, the expected trend in the dataset would be: for the same series, larger models generally perform better in all tasks across different context lengths. As shown in \cref{fig: verification}, this overall trend is observed, indicating that the dataset generation itself is reliable and can be used for evaluating long-context capabilities. For instance, compare to Llama 3.2-1B-Instruct, Llama 3.2-3B-Instruct gets higher average scores in each task. For more detailed results of models across various context lengths, refer to 
 \cref{Verification Appendix}.

\subsection{Verification of the effectiveness of the proposed metric}

Following the setting of \citet{lu2024controlled}, we compare two long-context enhancement methods, NTK and PI, using LongBench and \ourmethodNoColor. On \ourmethodNoColor, we evaluate performances with two metrics: score and \ourmetric($LC$). We include three evaluations to further validate the discriminative power and practical value of our proposed LongScore metric. These comparisons were chosen to reflect real-world modeling choices and align with community intuition: (1) NTK vs. PI on long-context tasks, (2) performance of LLaMA3-8B-Instruct under different RoPE theta ratios, and (3) Gemini-1.5 model variants like Gemini-1.5-Flash and Gemini-1.5-Pro from HEMLET benchmark. 

There are some reasons why we choose these three comparisons: (1) NTK and PI are chosen for comparison because it is well-established that NTK provides a more fine-grained extension of PI. (2) We choose LLaMA 3-8B-Instruct (8k claimed context length) with different RoPE theta ratios. Generally speaking, appropriately increasing the RoPE theta improves the model's long context capability (within a reasonable extent). (3) we choose Gemini-1.5-Flash and Gemini-1.5-Pro because they have an obvious difference in long-context ability.

On the LongBench tasks, both NTK and PI methods perform similarly, failing to provide a clear distinction. However, as shown in \cref{table:verification ntk pi}, on \ourmethodNoColor, the differences between NTK and PI became much more apparent across the selected tasks, effectively differentiating the two methods. Moreover, it is obvious that the differences of NTK and PI  measured by \ourmetric are greater than those measured by the traditional metric, showing that \ourmetric demonstrates a greater ability to highlight these differences compared to the traditional metric and a more effective tool for distinguishing long-context capabilities. 

In other pairwise comparison, LongScore readings show a much more pronounced difference compared to the original scoring metrics of the datasets, while the win-loss order remains consistent with our general understanding of a model or method's long context capability (NTK > PI, ratio=64 > ratio=1, Gemini-1.5-Pro > Gemini-1.5-Flash). These results highlight the discriminative power and effectiveness of our LongScore. 

\vspace{-1em}
\subsection{Experiments on frontier open-source LLMs }

This section introduces the experiments conducted using \ourmethodNoColor and the proposed metric, aimed at evaluating the long-context capabilities of various popular open-source large models.

We select four models, due to GPU resource limitations, as they can be used to generate outputs with a $256k$ context length. For each of the eight tasks, we generated 100 samples at each context length ($8k$, $16k$, $32k$, $64k$, $128k$) to obtain the scores, using the performance at $2k$, $4k$, and $6k$ as Base Ability. Finally, the average scores across all tasks are computed. \cref{table:our results on metrics} presents average results and the corresponding rankings. \cref{fig:results original} displays average scores at each context length. 

\begin{figure*}[t]
  \includegraphics[width=1\linewidth]{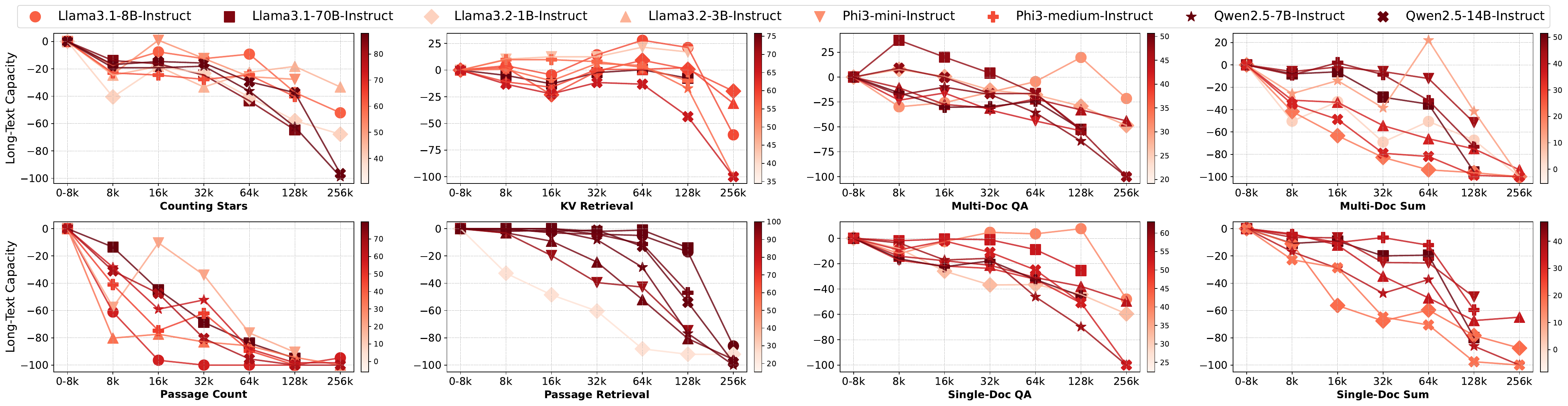}\vspace{-10pt}
  \caption {Results of eight open-source models on eight tasks are presented, with their scores calculated using \ourmetric metric. Each markrer represents a single model. The darker the color of the line, the stronger the base ability of the model.}
 \label{fig: results}
\end{figure*}
\begin{figure*}[t]
\includegraphics[width=1\linewidth]{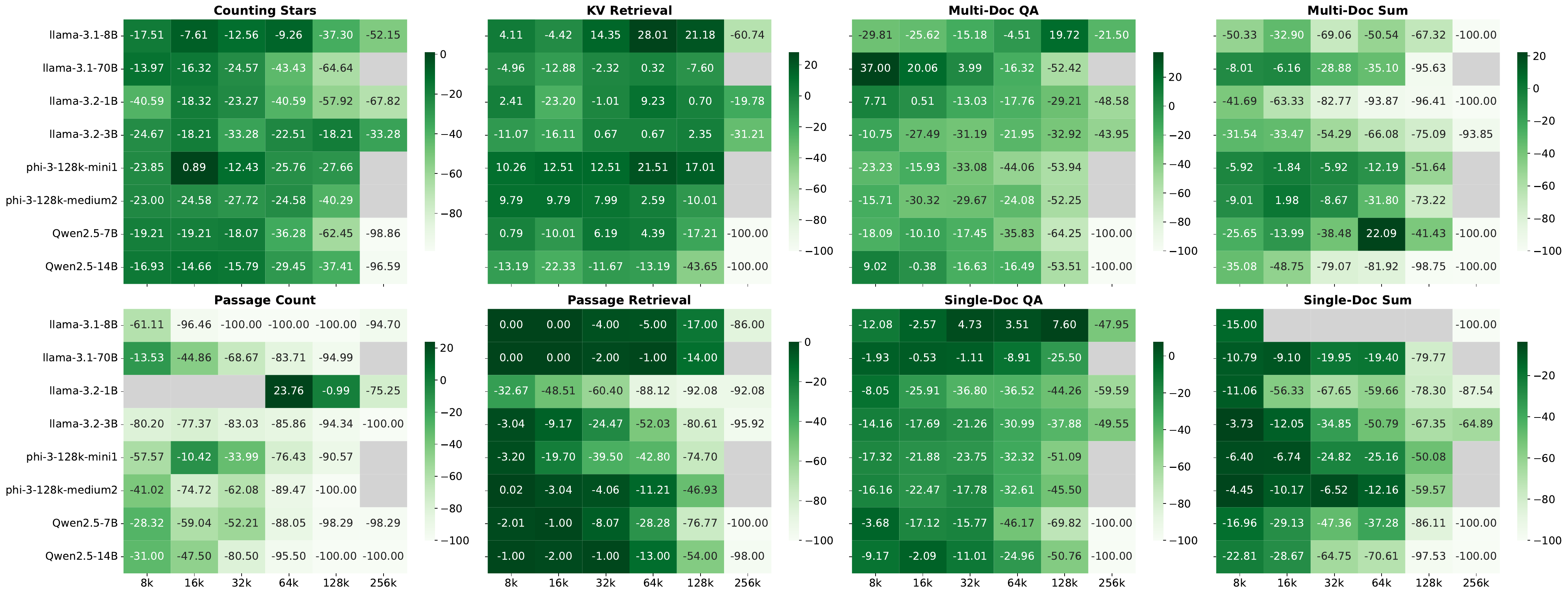}  
  \caption {Results of eight models on \ourmethodNoColor by using \ourmetric metric. The gray shading indicates either anomalous models' scores or cases where the model is unable to generate outputs for $256k$-long contexts.
  }\label{fig: results2}
\end{figure*}

Here we explain why we choose the appropriate context lengths (e.g. 2k, 4k, 6k) for measuring Base Ability. We evaluate  8 models spanning the \texttt{LLaMA 3.1}, \texttt{Phi-3}, and \texttt{Qwen 2.5} families. These models typically undergo pretraining with context lengths of 4K or 8K tokens before undergoing further continuous pretraining for long-context extension. Given this, we generalize that most models in our study have a pre-extension context window of either 4K or 8K. To probe their base reasoning ability, we evaluate performance under 2K, 4K, and 6K context lengths. These values are chosen to provide representative coverage of the model's original pretraining range without exceeding it, thereby offering a stable measure of Base Ability.

Interestingly, as shown in \cref{table:our results on metrics}, the rankings obtained by the traditional metric are almost identical to the rankings based on Base Ability. However, rankings using \ourmetric metric show a significant difference from Base Ability rankings, as demonstrated by models like Qwen 2.5-14B-Instruct and Qwen 2.5-7B-Instruct. From \cref{fig:results original}, it can be observed that while these two models have higher scores at shorter context lengths (e.g. 8k, 16k), their scores drop significantly at longer context lengths (128k, 256k). This indicates that current long-text evaluation metrics are heavily influenced by Base Ability, while 
\ourmetric(the metric proposed in this paper) separates base ability from long-context capability, providing a more accurate reflection of the model’s long-context performance. For comparisons of more open-source models on \ourmethodNoColor and their long-context capability evaluation, please refer to \cref{Open-source models on our proposed benchmark}.

We also present the results of eight models from four LLM family trees (Llama 3.1, Llama 3.2, Qwen 2.5 and Phi 3) on \ourmethodNoColor. The evaluation uses \ourmetric metric and the detailed results about each task are shown in \cref{fig: results} and \cref{fig: results2}.

Long-context ability is important in certain specialized domains such as healthcare and law. To this end, we additionally include several domain-specific long-context tasks, including Medical-Summary, MedOdyssey~\cite{fan2024medodysseymedicaldomainbenchmark}, and CaseSumm~\cite{heddaya2024casesummlargescaledatasetlongcontext}. We re-evaluate the performance of the \texttt{LLaMA 3.2-1B-Instruct} model with and without these datasets. The detailed results are shown in \cref{Results of models with and without domain-specific tasks}.

\subsection{Experiments on Ruler with different metrics}
\label{Results of Comparison of Different Methods for Enhancing Long-Text Capability}

We utilize data from Ruler~\cite{hsieh2024ruler}, using a $4k$-length context to represent the model’s base ability. The results are shown in \cref{table:ruler results on metrics2}, where we evaluate four models’ performance at different context lengths using both \ourmetric and the traditional metric. Compared to LLaMA 3.1 (70B), Yi (34B)~\cite{young2024yi} has a slightly lower overall score before reaching 128k context length, but at 128k, Yi (34B) performs significantly better. Similarly, compared to Phi3-medium (14B), LWM (7B) shows lower base ability and shorter text handling but clearly outperforms Phi3-medium at 128k. If ranking is based solely on scores, LLaMA 3.1 (70B) and Phi3-medium (14B) would be ranked higher than their counterparts, but this does not show their true long-context capabilities. By using \ourmetric, we correct this discrepancy.

\vspace{-0.5em}
\section{Related Works}
In this section, we review relevant prior research connected to our study. We summarize cutting-edge models known for their strong long-text processing capabilities, explore methods designed to enhance these abilities, and examine the benchmarks commonly used to assess long-text proficiency. Additionally, wwe discuss the limitations of existing benchmarks, not disentangling Base Ability from true long-context capabilities.


\textbf{Long-context language models.} 
Both open-source and closed-source state-of-the-art models now support extended context lengths of up to 128K tokens or more, including GPT-4~\cite{achiam2023gpt}, Gemini~\cite{team2024gemini}, Claude~\cite{caruccio2024claude}, LLaMA-3~\cite{dubey2024llama}, and Phi-3~\cite{abdin2024phi}. These models typically achieve long-context capabilities through a combination of improved pretraining and post-training techniques. For instance, many models adopt two-stage or continued pretraining pipelines, where an initial short context window (e.g., 4K or 8K) is later extended to longer lengths (e.g., 128K) using scalable attention mechanisms such as FlashAttention~\cite{dao2022flashattention} and optimized positional encoding schemes~\cite{li2021sequence,xiong2023effective,hsu2024liger}. This trend is well-documented in recent technical reports~\cite{yang2024qwen2, abdin2024phi, dubey2024llama}, which highlight how careful adjustments to training schedules, data distribution, and architecture design contribute to stable performance in extreme long-context settings. Nonetheless, despite these advancements, effectively evaluating and comparing the true reasoning ability of such models in long-context scenarios remains a significant challenge in the real situations and scenarios.

\textbf{Long context methods}. Many studies have explored methods to extend the context window length of models during fine-tuning, with some approaches even achieving this without fine-tuning. Techniques such as Position interpolation (PI)~\cite{chen2023extending}, NTK~\cite{peng2023ntk}, YaRN~\cite{peng2023yarn} and SelfExtend~\cite{jin2024llm} manipulate RoPE (Rotary Position Embedding)~\cite{su2024roformer} to do length extension. Other methods, including Retrievers~\cite{xu2023retrieval}, StreamingLLM~\cite{xiao2023efficient}, LM-Infinite~\cite{han2024lm}, Longlora~\cite{chen2023longlora}, Inf-LLM~\cite{xiao2024infllm} and Landmark~\cite{mohtashami2023landmark}, focus on designing new attention architectures or exploiting specific phenomena in attention mechanisms~\cite{sun2024massive} to achieve length extension. Additionally, some works~\cite{jiang2023longllmlingua, li2023compressing} focus on reducing length extension to length compression via a summarization step, where long contexts are compressed or summarized before being processed by the model.

\textbf{Long-context benchmarks}. LongBench~\cite{bai2023longbench} and L-Eval~\cite{an2023leval} are early benchmarks for evaluating long-context capabilities. Later benchmarks, such as $\infty$-Bench~\cite{zhang2024infintebench}, extended the context length of datasets further. Subsequently, synthetic task-related benchmarks like NIAH(Needle In A Haystack),  and Ruler~\cite{hsieh2024ruler} emerged, focusing not only on evaluating contextual capabilities but also on examining models’ sensitivity to the positional appearance of text. More recently, benchmarks such as HELMET~\cite{yen2024helmet} and LV-Eval~\cite{yuan2024lv} introduced controllable context lengths and LLM-based metrics. Building on them, this work further considers prior model knowledge, and introduces a novel metric. 

\vspace{-0.5em}
\section{Conclusion}
Our benchmark and metric address key shortcomings in current evaluation methodologies, such as the inability to isolate long-context reasoning from baseline performance and reliance on insufficiently representative tasks. By incorporating real-world data, diverse task types and difficulties, and a novel metric (\ourmetric), \ourmethodNoColor provides a robust platform to evaluate and compare LLMs across varying context lengths. This allows for a deeper understanding of how models handle extended contexts while minimizing the influence of prior knowledge or base abilities. As LLMs continue to evolve, the ability to rigorously assess their long-context reasoning will play a critical role in identifying bottlenecks and guiding the design of next-generation models. Our approach sets a new standard for assessing LLMs, paving the way for more robust innovations in long-context evaluation. Furthermore, it will provide an actionable insight for optimizing model architectures and training strategies to enhance long-context capabilities.

\section*{Limitations}
\noindent 
The proposed metric requires models to demonstrate relatively strong base ability on the task. If a model’s base ability is insufficient, subsequent evaluations of long-context capabilities may exhibit significant fluctuations, making it less effective for comparing models’ long-context performance. Besides, when constructing the benchmark, it is necessary to select articles of varying lengths to assemble into noisy contexts. For shorter target lengths, such as 2k tokens, the selected articles should also have shorter lengths — preferably less than 1k tokens — to ensure the context can be formed with two or more documents. Therefore, it is essential to collect texts of diverse lengths, particularly shorter ones, to enable effective assembly of the desired contexts.

\section*{Acknowledgements}

\noindent This research was partially supported by NSF Awards OAC-2117439. Further, this work made use of the High Performance Computing Resource in the Core Facility for Advanced Research Computing at Case Western Reserve University (CWRU). We give special thanks to the CWRU HPC team for their prompt and professional help and maintenance. The views and conclusions in this paper are those of the authors and do not represent the views of any funding or supporting agencies.

\bibliography{ref}
\clearpage
\appendix
\section{Appendix}
\label{sec:appendix}
\subsection{Results of models’ long-text enhancement methods on Longbench}

\begin{figure*}[htbp]
    \label{fig: performances on three LongBench tasks}
    \centering
    \includegraphics[width=\linewidth]{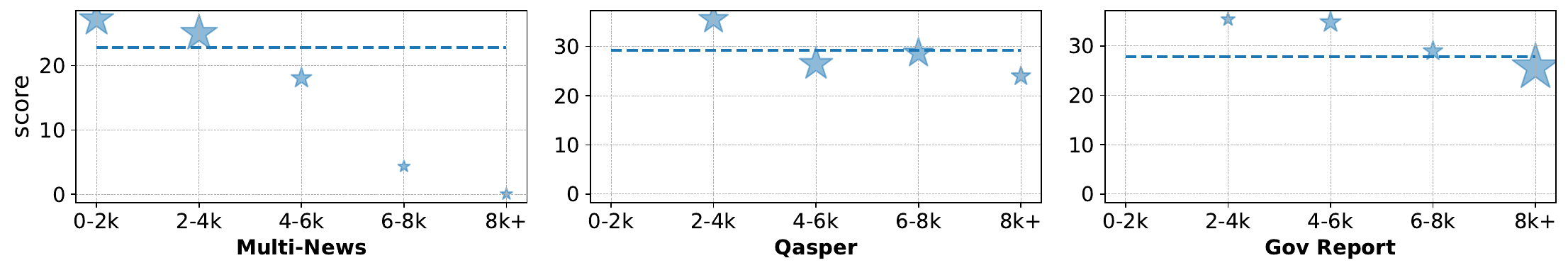}
    \caption{Illustration of NTK's performances on three LongBench tasks.}
    \vspace{1em}
    \includegraphics[width=\linewidth]{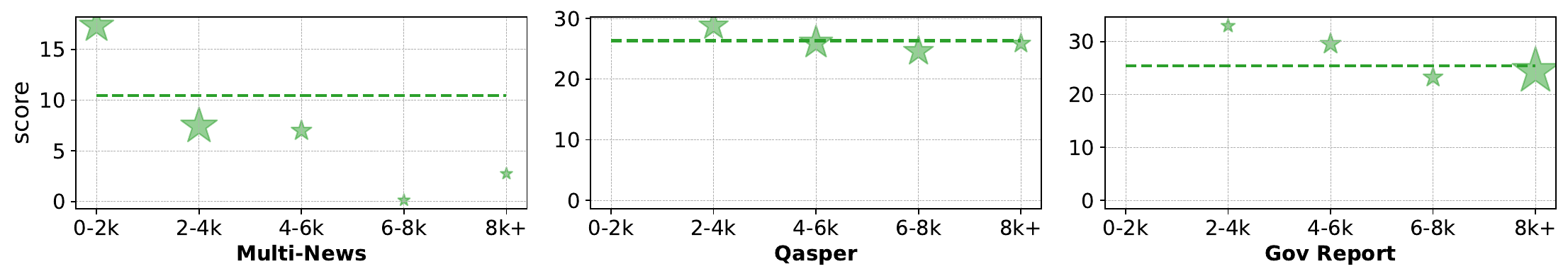}
    \caption{Illustration of PI's performances on three LongBench tasks.}
    \vspace{1em}
     \includegraphics[width=\linewidth]{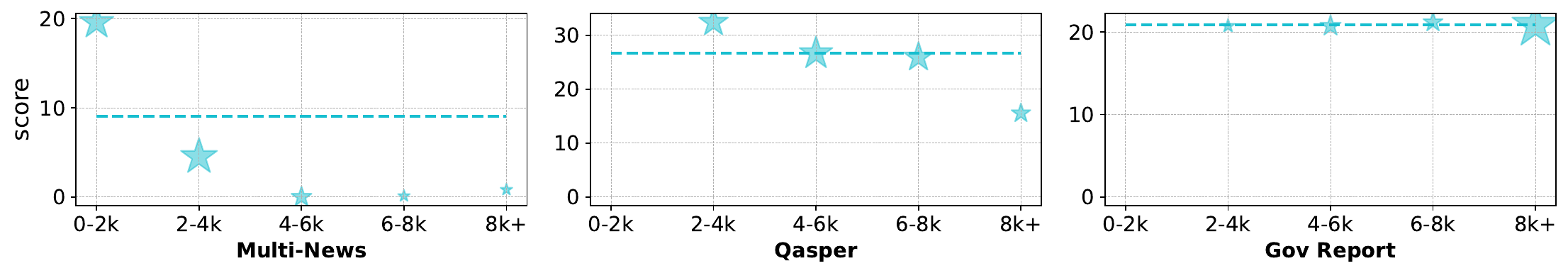}
    \caption{Illustration of YaRN's performances on three LongBench tasks.}
    \vspace{1em}
     \includegraphics[width=\linewidth]{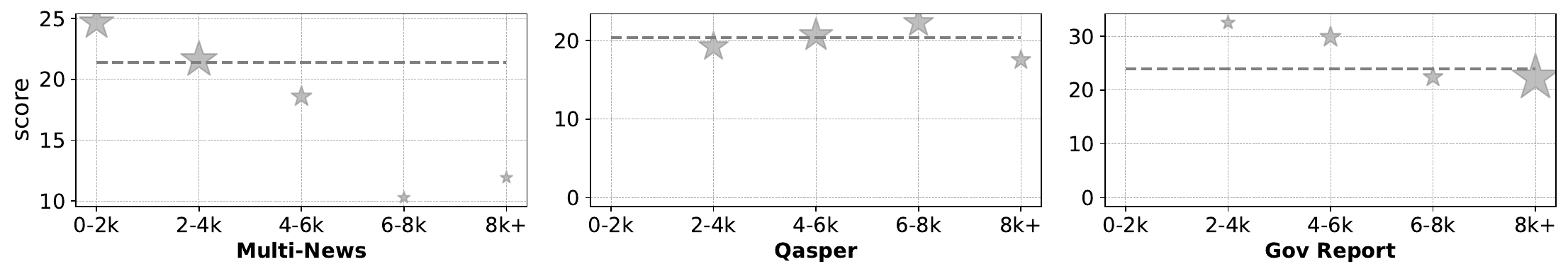}
    \caption{Illustration of Longlora's performances on three LongBench tasks.}
    \vspace{1em}
    \includegraphics[width=\linewidth]{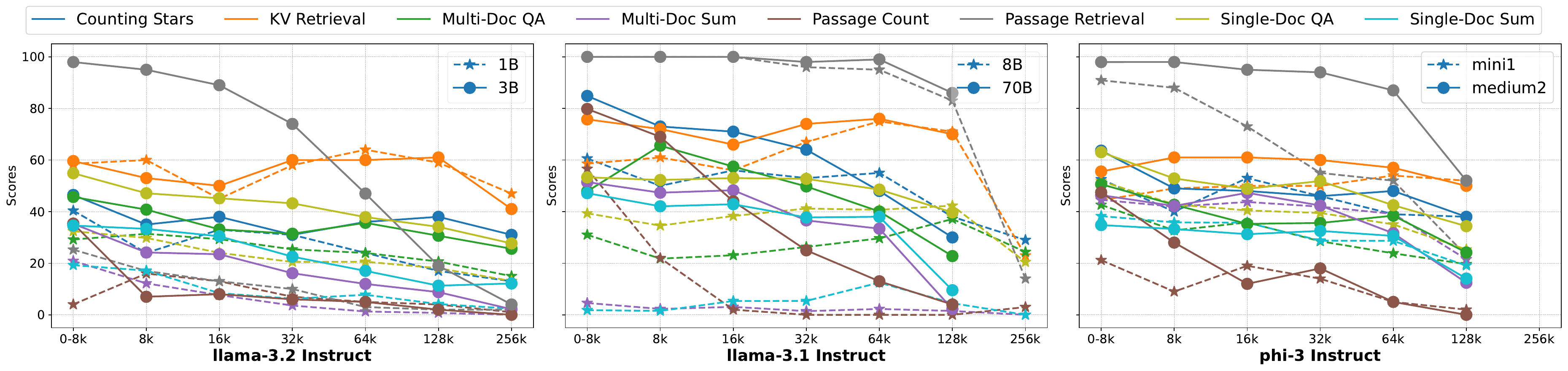}  
      \caption {Verification the reliability of \ourmethodNoColor: results of two models of different sizes from the same LM family tree, showcasing their scores in different tasks across various context lengths. One color represents a specific task, with solid lines indicating larger models and dashed lines representing smaller models. The results of different LMs from the same LM family tree basically validate the general trend: the larger model tend to get a higher score while the score decreases as the context length increases.}
     \label{fig: verification line}
\end{figure*}

\label{Results of models’ long-text enhancement methods on Longbench}

These section introduces four long-context enhancement method's performances on three LongBench tasks. The colored dashed lines represent the average score of each model on the corresponding task. The size of the markers corresponds to the proportion of each text length within the entire dataset. The larger the marker, the higher the proportion. The results exhibit significant variation across tasks of different lengths within the same dataset. All results are in \cref{fig: performances on three LongBench tasks}.

\subsection{Details about how to construct each task}

\label{sec:construction}
\textbf{KV Retrieval}. This task primarily evaluates the model’s ability to extract critical information while ignoring irrelevant content and noisy information. (1) Context Construction: Three pairs of key-value ($k_1$, $v_1$; $k_2$, $v_2$; $k_3$, $v_3$) are generated using UUIDs. The value of the previous pair serves as the key for the subsequent pair ($v_1$ = $k_2$; $v_2$ = $k_3$). These key-value pairs are randomly inserted into different noisy contexts. The noise introduces irrelevant or distracting information, simulating real-world challenges. (2) Question Setup: The question asks the model to identify the value corresponding to a specific key. (3) Evaluation Metric: The task is evaluated using accuracy (\text{Acc}). If the model correctly identifies the value associated with the queried key, its accuracy score is incremented by one. 

\textbf{Counting Stars}. Following \cite{song2024countingstar}
, this task assesses the model’s ability to extract critical information across multiple documents, maintain the correct sequence when aggregating information and resist distractions from misleading or altered options. (1) Context Construction: Four noisy context passages are selected from all noisy context passages and each passage is appended with a sentence in the format: \textit{The little penguin counted $N$ \ding{72}}, where $N$ represents a specific number of stars counted in that passage. (2) Question Setup: The model is tasked with identifying the sequence of star counts in the order of sentence appearance, like [38, 10, 90, 42]. The task provides multiple-choice options, including the correct sequence and several distractors. Distractors are generated by swapping numbers, modifying values, or changing the order to increase difficulty. (3) Evaluation Metric: The task is evaluated using accuracy (\text{Acc}). If the model selects the correct sequence, its accuracy score is incremented by one.

\textbf{Passage Retrieval}. By focusing on comprehension and recognition, this task challenges the model’s ability to extract and correlate key information in a multi-document setting. (1) Context Construction: A single data sample comprises multiple articles, each sourced from a distinct domain. These articles are concatenated to form the context. (2) Question Setup: The model is provided with the summary of one specific article from the context. The task is to identify which article in the context corresponds to the given summary. (3) Evaluation Metric: The task is evaluated using accuracy (\text{Acc}). If the model correctly identifies the article corresponding to the summary, its accuracy score is incremented by one.

\textbf{Passage Count}. The task assesses a model’s ability to understand and integrate global key information by determining the number of unique articles within a multi-article context. (1) Context Construction: Each data sample comprises multiple articles sourced from different domains. Some articles are repeated multiple times within the context to add redundancy and complexity. (2) Question Setup: The model is tasked with identifying the total number of unique (non-repeated) articles in the context. (3) Evaluation Metric: The task is evaluated using accuracy (\text{Acc}). If the model correctly identifies the count of unique articles, its accuracy score is incremented by one.

\textbf{Single-Doc QA}. The task evaluates a model’s ability to answer questions specific to a single article within a multi-article context. (1) Context Construction: Each data sample consists of multiple articles from different domains. A specific question is posed about one particular article within the context. (2) Evaluation Metric: The model’s answers are assessed using another large language model (like GPT-4o-mini). Evaluation is based on two dimensions: Fluency is scored on a 3-point scale (0, 1, 2), evaluating the coherence and readability of the answer. Correctness is scored on a 4-point scale (0, 1, 2, 3), assessing the factual accuracy of the response in relation to the context. The final score is calculated as the product of the Fluency and Correctness scores: $\text{Final Score} = \text{Fluency} \times \text{Correctness}$ (3) Prior Knowledge Filtering: To filter out the model’s prior knowledge, we introduce a filtering process. In a no-context scenario, if the model’s response score exceeds a certain threshold, it indicates that the model is relying on prior knowledge. In such cases, the data is excluded from the statistical analysis.

\textbf{Multi-Doc QA}. The task evaluates a model’s ability to integrate information from multiple articles and provide coherent, accurate answers to questions that require a global understanding of the context. (1) Context Construction: Each data sample contains multiple articles from different domains. The question posed requires the model to synthesize information across multiple articles to generate the correct answer. (2) Evaluation Metric: Similar to the Single-Doc QA task, the model’s answers are evaluated using another large language model and evaluated by the same dimensions. (3) Prior Knowledge Filtering is similar to the Single-Doc QA task. 

\textbf{Single-Doc Sum}. The task evaluates a model’s ability to generate concise and accurate summaries for a specific article within a multi-article context. (1) Context Construction: Each data sample consists of multiple articles from different domains. (2) Question Setup: The model is tasked with summarizing the content of one specific article from the context. (3) Evaluation Metric: The generated summary is assessed by another large language model. Two scoring dimensions are considered: Fluency evaluates the coherence and readability of the summary and is scored on a 2-point scale: 0 (poor fluency), 1 (good fluency). Precision measures the relevance of the summary by comparing each sentence in the model’s output to the reference summary. and is calculated as $ \text{Precision} = \frac{\text{Number of relevant sentences}}{\text{Total number of sentences in the summary}} $. The final score is the product of these two dimensions: $ \text{Final Score} = \text{Fluency} \times \text{Precision} $. By requiring accurate and readable summaries, this task emphasizes the model’s capacity for effective information synthesis and integration.

\textbf{Multi-Doc Sum}. The task evaluates a model’s ability to integrate information from multiple articles and produce a coherent and accurate summary of their shared content. (1) Context Construction: Each data sample consists of multiple articles from different domains. (2) Question Setup: The model is tasked with summarizing the relevant content from all provided articles. (3) Evaluation Metric: Similar to the Single-Doc Sum task, the model’s answers are evaluated using another large language model and evaluated by the same dimensions. By requiring effective summarization of multi-document content, this task highlights the model’s ability to synthesize and generalize information across diverse sources.

\subsection{Prompts used in each task}
This section presents the prompts used in each task. Here, \textit{\{context\}} represents the entire context constructed from articles in the noisy context sources and real context sources. \textit{\{input\}} represents the question for the task, and \textit{\{instruction\}} represents the model-specific instructions. For example, in Single-Doc QA, the instruction might be “Answer the question related to Passage 1”, indicating that the question is specifically based on Passage 1.

\textbf{KV Retrieval}. \textit{There are some passages below sourced from many different fields.\textbackslash n\textbackslash n \{context\} \textbackslash n\textbackslash n Given several key-value pairs in these passages, you need to find the value of the key. Read the question related with these key-value pairs and give the correct answer. \{input\}}

\textbf{Counting Stars}.  \textit{There are some passages below sourced from many different fields.\textbackslash n \textbackslash n \{context\} \textbackslash n\textbackslash n Only output the results without any explanation. Read the following question and give the correct answer: \{input\} \textbackslash n The final answer is: }

\textbf{Passage Retrieval}. \textit{Here are some passages from many different fields, along with an summarization. Please determine which passage the summarization is from.\textbackslash n \textbackslash n \{context\} \textbackslash n \textbackslash n The following is a summarization.\textbackslash n\textbackslash n \{input\} \textbackslash n \textbackslash n Please enter the number of the passage that the summarization is from. The answer format must be like "Passage 1", "Passage 2", etc. \textbackslash n\textbackslash n The answer is Passage  }

\textbf{Passage Count}. \textit{There are some paragraphs below sourced from many different fields. Some of them may be duplicates. Please carefully read these paragraphs and determine how many unique paragraphs there are after removing duplicates. In other words, how many non-repeating paragraphs are there in total? \textbackslash n\textbackslash n \{context\} \textbackslash n\textbackslash n Please enter the final count of unique paragraphs after removing duplicates. The output format should only contain the number, such as 1, 2, 3, and so on.\textbackslash n\textbackslash n The final answer is: }

\textbf{Single-Doc QA}. \textit{Answer the question based on the given passages. Only give me the answer and do not output any other words.\textbackslash n \textbackslash n The following are given passages and these passages are from many different fields.\textbackslash n \textbackslash n \{context\} \textbackslash n \textbackslash n Answer the question based on the given passages following the instruction: \textbackslash n \{instruction\} \textbackslash n \textbackslash n Question: \{input\} \textbackslash n Only give me the answer and do not output any other words. Answer: \textbackslash n",}

\textbf{Multi-Doc QA}. \textit{Answer the question based on the given passages. Only give me the answer and do not output any other words.\textbackslash n \textbackslash n The following are given passages and these passages are from many different fields.\textbackslash n \textbackslash n \{context\} \textbackslash n \textbackslash n Answer the question based on the given passages following the instruction: \textbackslash n \{instruction\} \textbackslash n \textbackslash n Question: \{input\} \textbackslash n Only give me the answer and do not output any other words. Answer: \textbackslash n }

\textbf{Single-Doc Sum}. \textit{You are given several passages as follows, but not all of them need to be summarized. \textbackslash n \textbackslash n \{context\} \textbackslash n \textbackslash n Please follow these instructions: \textbackslash n 1.\{input\} \textbackslash n 2.Ignore and do not summarize any passages not listed above. \textbackslash n 3.For the selected passages, the summary should include: the main arguments or conclusions of each article, the key evidence or supporting data presented and any unique or innovative points made in the passages. \textbackslash n 4.The summary should be concise, focusing only on the most important information from the passages. Now, please generate the summary for the specified passage, following the instructions carefully. \textbackslash n Summary: \textbackslash n}

\textbf{Multi-Doc Sum}. \textit{You are given several passages as follows, but not all of them need to be summarized.\textbackslash n \textbackslash n \{context\} \textbackslash n \textbackslash n Please follow these instructions:\textbackslash n 1.\{input\} \textbackslash n 2.Ignore and do not summarize any passages not listed above. \textbackslash n 3.All the selected passages should be summarized into a few short sentences and do not summarize each selected passages separately. The summary should include: the main arguments or conclusions of each article, the key evidence or supporting data presented and any unique or innovative points made in the passages. \textbackslash n 4.The summary should be concise, focusing only on the most important information from the passages. Now, please combine and summarize the main ideas from the selected relevant passages into one cohesive summary, following the instructions carefully.\textbackslash n \textbackslash n Summary: \textbackslash n }

\subsection{Further verification of the reliability of the proposed benchmark}

\label{Verification Appendix}
To further verify the reliability of the generated dataset, we evaluate three model families (Llama 3.2, Llama 3.1, and Phi 3), selecting two different model sizes from each family. Given that these models are from the same series but vary in size, the expected trends on the dataset are as follows: (1) Model Size Effect: Larger models should generally achieve higher scores compared to smaller models within the same series. (2) Text Length Effect: As the text length increases, the performance scores should decrease across all models. As shown in \cref{fig: verification line}, the results basically follow these expected trends: larger models tend to score higher, and performance decreases as text length increases. This consistent pattern indicates that the dataset generation process is accurate and reliably.

\begin{table}
  \centering
  \caption{
    Results of the average performance of five models across all tasks on \ourmethodNoColor. \textbf{Base Ability} represents the model’s score within lengths of $2k$, $4k$ and $6k$ \textbf{$Avg$ score} represents the average of score across lengths including $8k$, $16k$, $32k$, $64k$ and $128k$. \textbf{$Avg$ LC} represents the average of score by using our proposed metric. $57.4_{(1)}$ indicates that the current model has a score of 57.4 at the given context length, with a ranking of 1. Claimed Length refers to the maximum context length that the model claims to support. Qwen 2.5-14B and Qwen 2.5-7B use YaRN to extend their context length to 128k. so, the original context length is specified in Claimed Length.
    }\vspace{-10pt}
    
  \resizebox{\columnwidth}{!}{
  \begin{tabular}{c|c|c|cc}
     \toprule
  \multirow{2}{1.5em}{\centering  \textbf{model}} & \multirow{2}{4em}{\centering Claimed Length} & \multirow{2}{3em}{\centering  \textbf{Base Ability}} & \multirow{2}{2em}{\centering $Avg$ \textbf{score}}& \multirow{2}{2em}{\centering $Avg$ \textbf{LC}} \\
    \\
    \midrule

    llama-3.1-70B-Instruct & 128K & $67.5_{(1)}$
    & $52.55_{(1)}$ & $-22.18_{(2)}$ \\
    Qwen2.5-14B-Instruct &  32K & $59.1_{(2)}$
    & $40.77_{(3)}$ & $-31.12_{(7)}$ \\
    Phi-3-128k-medium & 128K & $57.4_{(3)}$
    & $43.28_{(2)}$ & $-24.65_{(4)}$ \\
    Qwen2.5-7B-Instruct &  32K  & $57.4_{(4)}$
    & $39.80_{(4)}$ & $-30.69_{(6)}$ \\
    Llama3.2-3B-Instruct  & 128K   &  $51.2_{(8)}$
    & $34.81_{(7)}$ & $-32.06_{(8)}$ \\
    Phi-3-128k-mini& 128K & $48.2_{(6)}$
    & $36.78_{(5)}$ & $-23.85_{(3)}$ \\
    Llama-3.1-8B-Instruct  & 128K   & $44.0_{(7)}$
    & $36.37_{(6)}$ & $-17.46_{(1)}$ \\
    Llama3.2-1B-Instruct  & 128K   & $28.7_{(8)}$
    & $20.45_{(8)}$ & $-28.88_{(5)}$ \\
    
   \bottomrule
  \end{tabular}
  }
  \label{table:our all results on metrics}
\end{table}

\begin{figure}[t]
  \includegraphics[width=\columnwidth]{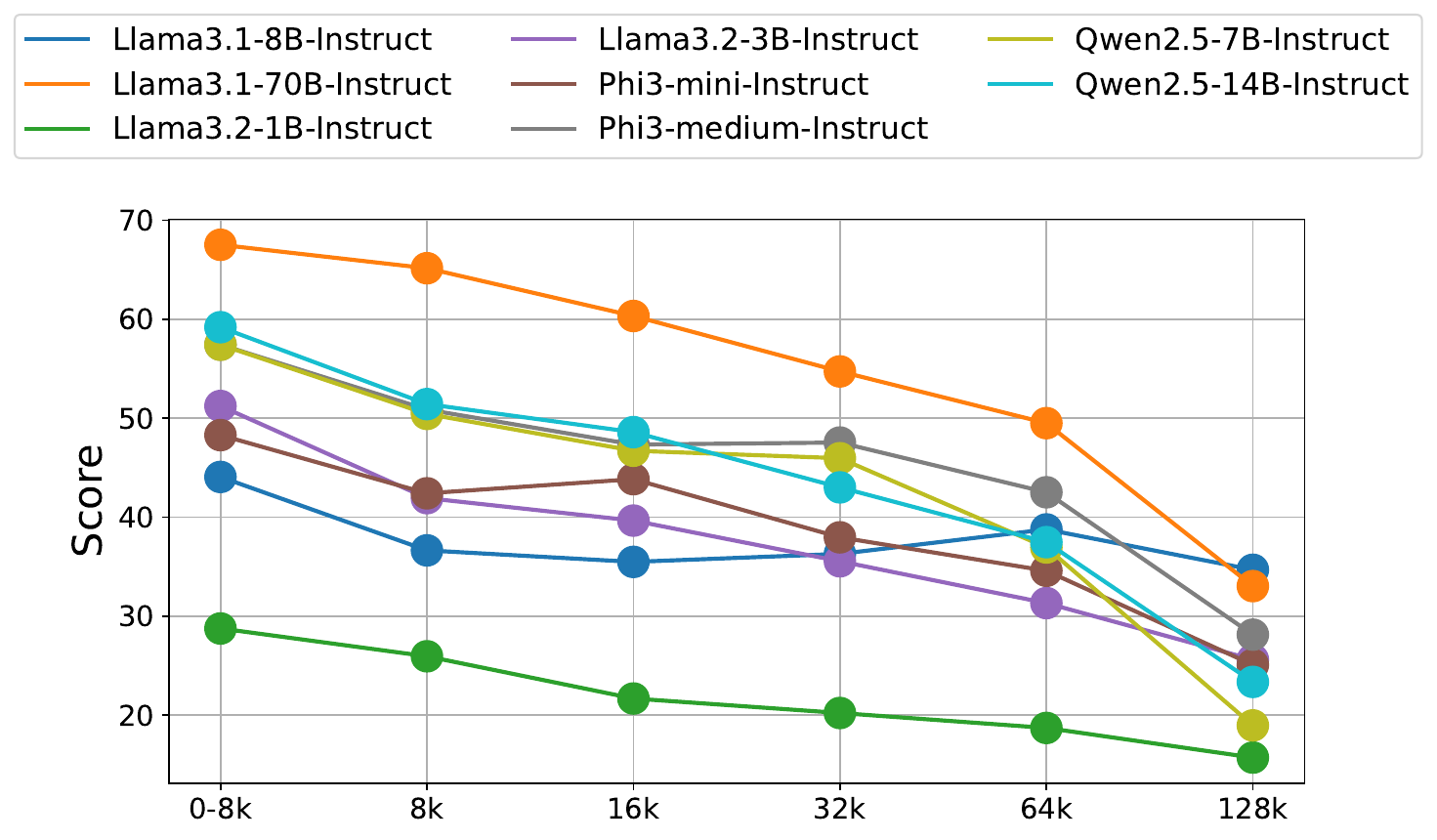}
  \caption{Results of eight open-source models on all tasks in \ourmethodNoColor, showing their scores at different context lengths. }
  \label{fig:all results original}
\end{figure}

\begin{table*}[ht]
\centering
\caption{
Performance of \texttt{LLaMA 3.2-1B-Instruct} with and without domain-specific tasks.
We report scores across different context lengths and two average metrics: overall average and average on long contexts (32k+). Adding healthcare and law tasks leads to a slight drop in average long-context performance.
}
\vspace{-10pt}
\label{tab:domain-tasks}
\small{
\resizebox{\textwidth}{!}{
\begin{tabular}{l|cccccc|cc}
\toprule
\textbf{Benchmark} & \textbf{base} & \textbf{8k} & \textbf{16k} & \textbf{32k} & \textbf{64k} & \textbf{128k} & \textbf{avg(score)} & \textbf{avg(LongScore)} \\
\midrule
original & 24.41 & 22.42 & 20.55 & 18.54 & 17.92 & 15.44 & 18.97 & -22.27 \\
original + healthcare \& law & 24.58 & 21.97 & 18.49 & 15.77 & 16.64 & 12.83 & 17.14 & -30.27 \\
\bottomrule
\end{tabular}
}
}
\end{table*}

\label{Results of Different
Open-source models}

\subsection{Results of different
Open-source models on our proposed benchmark}
\label{Open-source models on our proposed benchmark}

This section first introduces the experiments conducted using \ourmethodNoColor and the proposed metric, aimed at evaluating the long-context capabilities of various popular open-source large models.

We select eight open-source models. For each of the eight tasks, we generated 100 samples at each context length ($8k$, $16k$, $32k$, $64k$ and $128k$) to obtain the scores. The model’s Long-context Capability was then calculated, using the performance at $2k$, $4k$, and $6k$ as the base ability. Finally, the average scores across all tasks for the five models are computed. \cref{table:our all results on metrics} presents the final average results and the corresponding rankings of the five models. \cref{fig:all results original} displays the average scores for all tasks at each context length for the five models. 

\subsection{Results of models with and without domain-specific tasks}
\label{Results of models with and without domain-specific tasks}

We have added long text datasets from the recommended domains (law and healthcare) to enhance the comprehensiveness of our benchmark. Evaluating the capability of LLMs to handle such domain-specific scenarios is indeed a crucial need.

Specifically, we mix up CaseSumm, MedOdyssey, and Medical Summary into our original dataet. We reevaluate the performance of the LLaMA 3.2 1B-Instruct model with and without such datasets.

As is shown in \cref{tab:domain-tasks}, incorporating healthcare and law-focused domain-specific data leads to a slight performance decline in long text scenarios, likely because the model lacks comprehensive knowledge in these specialized fields. However, the overall trend is steady. We plan to incorporate this additional evaluation to our updated manuscript and add more discussion regarding domain-specific long context evalutions.

\end{document}